\documentclass[runningheads]{llncs}
\usepackage{graphicx}
\usepackage{hyperref}
\usepackage{tabularx}
\usepackage[export]{adjustbox}
\begin{document}

\title{Real-World Deployment and Evaluation of Kwame for Science, An AI Teaching Assistant for Science Education in West Africa}

\titlerunning{Real-World Deployment and Evaluation of Kwame for Science}


\author{George Boateng\inst{1,2} \and
Samuel John\inst{1}\and
Samuel Boateng\inst{1}\and
Philemon Badu \inst{1}\and
Patrick Agyeman-Budu \inst{1}\and
Victor Kumbol \inst{1}}
\authorrunning{Boateng et al.}

\institute{Kwame AI Inc.\\ 
\and
ETH Zurich, Switzerland}



\maketitle   










\begin{abstract}
Africa has a high student-to-teacher ratio which limits students’ access to teachers for learning support such as educational question answering. In this work, we extended Kwame, a bilingual AI teaching assistant for coding education, adapted it for science education, and deployed it as a web app. Kwame for Science provides passages from well-curated knowledge sources and related past national exam questions as answers to questions from students based on the Integrated Science subject of the West African Senior Secondary Certificate Examination (WASSCE). Furthermore, students can view past national exam questions along with their answers and filter by year, question type, and topics that were automatically categorized by a topic detection model which we developed (91\% unweighted average recall). We deployed Kwame for Science in the real world over 8 months and had 750 users across 32 countries (15 in Africa) and 1.5K questions asked. Our evaluation showed an 87.2\% top 3 accuracy (n=109 questions) implying that Kwame for Science has a high chance of giving at least one useful answer among the 3  displayed. We categorized the reasons the model incorrectly answered questions to provide insights for future improvements. We also share challenges and lessons with the development, deployment, and human-computer interaction component of such a tool to enable other researchers to deploy similar tools. With a first-of-its-kind tool within the African context, Kwame for Science has the potential to enable the delivery of scalable, cost-effective, and quality remote education to millions of people across Africa. 


\end{abstract}



\section{Introduction}
The COVID-19 pandemic has exacerbated the already poor educational experiences of millions of students in Africa who were grappling with educational challenges like poor access to computers, the internet, and teachers. In 2018, the average student-teacher ratio in Sub-Saharan Africa was 35:1 which is higher compared to 14:1  in Europe \cite{UNESCO2020}. In this context, students struggle to get answers to their questions. Hence, offering quick and accurate answers, outside of the classroom, could improve their overall learning experience. However, it is difficult to scale this support with human teachers.

In 2020, Boateng developed Kwame \cite{boateng2021b}, a bilingual AI teaching assistant that provides answers to students’ coding questions in English and French for SuaCode, a smartphone-based online coding course \cite{boateng2019,boateng2021}. Kwame is a deep learning-based question-answering system that finds the paragraph most semantically similar to the question via cosine similarity with a Sentence-BERT (SBERT) model  \cite{reimers2020}. 

We extended Kwame to work for science education in West Africa — Kwame for Science — and deployed it as a web app. Kwame for Science \footnote{\href{http://ed.kwame.ai/}{http://ed.kwame.ai/}} provides passages from well-curated knowledge sources and related past national exam questions as answers to questions from students based on the Integrated Science subject of the West African Senior Secondary Certificate Examination (WASSCE). This subject is a core subject that covers various aspects of science such as biology, chemistry, physics, earth science, and agricultural science. It is mandatory for senior high school students in some West African Education Council (WAEC) member countries such as Ghana and The Gambia. Furthermore, students can view past national exam questions along with their answers and filter by year, question type (objectives, theory, and practicals), and syllabus topics (that were automatically categorized). We deployed and evaluated Kwame for Science in the real world over 8 months. This work builds upon our prior work-in-progress paper which only described the question-answering aspect of Kwame for Science and provided preliminary results for its deployment over 2.5 weeks \cite{boateng2022}.

Our key contribution is the development of a system for (1) science question answering using a state-of-the-art neural retriever — SBERT and (2) automatically categorizing syllabus topics using SBERT and a support vector machine, both for the unique context of Science education in West Africa along with its deployment and evaluation as an AI teaching assistant in an extended, real-world, large scale context (over 700 people primarily in West Africa over 8 months).

\section{Related Work: Virtual Teaching Assistants}
\label{sec:related_work}
There are virtual teaching assistants (TA) such as Jill Watson \cite{goel2016,goel2020}, Rexy \cite{benedetto2019}, a physics course TA \cite{zylich2020}, and Curio SmartChat (for K-12 science) \cite{raamadhurai2019}. The first work on virtual TAs was done by Professor Ashok Goel at the Georgia Institute of Technology in 2016. His team built Jill Watson, an IBM Watson-powered virtual TA to answer questions on course logistics in an online version of an Artificial Intelligence course for master’s students \cite{goel2016}. It used question-answer pairs from past course forums. Given a question, the system finds the closest related question and returns its answer if the confidence level is above a certain threshold. Since then, various versions of Jill Watson have been developed to perform various functions (1) Jill Watson Q \& A (revision of the original Jill Watson) now answers questions using the class syllabi rather than QA pairs. (2) Jill Watson SA gives personalized welcome messages to students when they introduce themselves in the course forum after joining and helps create online communities. (3) Agent Smith aids in creating course-specific Jills using the course' syllabus \cite{goel2020}. Jill’s revision now uses a 2-step process. The first step uses commercial machine-learning classifiers such as Watson and AutoML to classify a sentence into general categories. Then the next step uses their proprietary knowledge-based classifier to extract specific details and gives an answer from  Jill's knowledge base using an ontological representation of the class syllabi that they developed. Also, the responses pass through a personality module that makes Jill sound more human-like. Jill has answered thousands of questions, in one online and various blended classes with over 4,000 students over the years.

Similar work has been done by other researchers who built a platform, Rexy for creating virtual TAs for various courses also built on top of IBM Watson \cite{benedetto2019}. The authors described Rexy, an application they built that can be used to build virtual teaching assistants for different courses. They also presented a preliminary evaluation of one such virtual teaching assistant in a user study. The system has a messaging component (such as Slack), an application layer for processing the request, a database for retrieving the answers, and an NLP component built on top of IBM Watson Assistant which is trained with question-answer pairs. They use intents (the task students want to be done) and entities (the interaction context). The intents (71 of those, e.g., exam date) are defined in Rexy but the entities have to be specified by instructors as they are course-specific. For each question, a confidence score is determined and if it is below a threshold, the question is sent to a human TA to answer and the answer is forwarded to the student. The authors implemented an application using Rexy and deployed it in an in-person course (about recommender systems) with 107 students. After the course, the authors reviewed the conversation log and identified two areas of requests (1) course logistics (e.g., date and time of exams) and (2) course content (e.g., definitions and examples about lecture topics).

In the work by Zylich et al. \cite{zylich2020}, the authors developed a question-answering method based on neural networks whose goal was to answer logistical questions in an online introductory physics course. Their approach entailed the retrieval of relevant text paragraphs from course materials using TF-IDF and extracting the answer from it using an RNN to answer 172 logistical questions from a past physics online course forum. They used as data the PhysicsForum dataset from a previous physics online course consisting of past Piazza posts (2004 posts, 802 questions, 172 logistical questions eg. due dates, exam location, grading policy)  and course materials (288 docs): syllabus, sections course textbook, lecture slides, emails, and forum posts. They used a two-method approach similar to Chen et al. \cite{chen2017}. The first is a document retriever based on DrQA and uses TF-IDF. It gets the 5 top-ranked documents. The second is an answer extractor which uses an RNN to find the start and end sequence to use as the answer. The RNN is trained on the SQuAD dataset. They also used the BERT-based model to determine if a question is answerable based on the question-document pair. The model is fine-tuned on SQUAD 2.0 and Natural Questions which had human-generated answerability scores. They also developed an RNN-based model to predict the time that the post will get a response using information like BERT embedding of the post, the time between the last and current post, and the post type. They evaluated their approach separately for document retrieval (dr) and answer extraction (ae). They used the metrics top 1, 3, and 5 corresponding to the accuracies when the correct answer is in the top 1, 3, or 5 returned answers. They compared their system with DrQA for document retrieval and Human and IBM Watson Assistant (which was provided with the top 15 question-answer pairs). They did an ablation study for document retrieval, splitting the documents into paragraphs and using normalized cosine similarity, They assessed the use of the answerability classifier and the next post-time prediction. They also applied their system to answering 18 factual course questions using document retrieval without extracting an answer and had top 1, 3, and 5 accuracies of 44.4\%, 88.9\%, and 88.9\% respectively.

Raamadhurai et al. developed Curio SmartChat, an automated question-answering system for K12 learning (middle school science topics) used by 20K students who asked over 100K questions \cite{raamadhurai2019}.  Their system consists of a QA engine that retrieves answers, a content library that contains the curriculum-relevant text content and metadata (figures) and media,  and a web app chat interface where students type their questions. When a student types a question, the system retrieves an answer. If it does not understand the question, it offers possible closely related questions from a bank of questions to disambiguate. If it does not find any recommended answer, it responds with small talk such as greetings. The content library contains text documents and quick definitions based on the curriculum. It also contains questions and answers tagged according to three levels of Bloom’s  Taxonomy: Knowledge, Understanding, and Application. They trained their custom taggers for intent and entity extraction by extending SpaCy taggers. The intent classifier decides which service will provide the response, small talk or content library, while the entity extractor will retrieve the entities the user is interested in. For example, if the query is “What is photosynthesis?”, then the tagged JSON would look like {“intent”: “content”, “entity”: [“photosynthesis”]}. There could be more than 1 entity but only 1 intent. They used the Universal Sentence Encoder model to encode the question and then compare it to vectors of the content via hashmap lookups. If the confidence is low, then it recommends similar questions. If the confidence is even lower, then it responds with small talk. Deployment was done with Tensorflow and Docker. Since launching the service, their system has served over 100,000 questions from around  20,000 students, mostly 13-16 years old. They mentioned that their system is mostly accurate once an answer is given because of their confidence checks but no robust evaluation was provided. For the case where similar questions were recommended or small talk responses were used, they sampled 200 random queries and 2 people individually rated them. They rated the response as valid if one of the recommendations was valid: 65\% were valid, 20 questions did not have available content to answer, 107 did have and 60 of the queries were not clear. Students had several spelling mistakes posing challenges for the system.

These works are focused on answering logistical questions, except Curio SmartChat whereas ours is focused on answering content-based questions which are more challenging. Compared to Curio SmartChat which is the closest work to ours, our work uses a state-of-the-art language model (SBERT) relative to theirs (Universal Sentence Encoder). Additionally, they did not provide an evaluation of the accuracy of their system after real-world deployment using ratings from users which we do. Furthermore, our work is the first to be developed and deployed in the context of high school science education in West Africa.

\section{Background: Question Answering (QA)}
\label{sec:background}
Within the domain of question answering, there are two main paradigms: Extractive QA and Generative QA. Extractive QA entails machine reading, which refers to selecting a span of text in a paragraph (called the context) that directly answers a question. To obtain the context, it can be directly provided such as in the task described for the SQuAD dataset \cite{rajpurkar2016}. Also, the most relevant passage(s) can be retrieved from various documents, a process called retrieval. This approach was used by Chen et al. \cite{chen2017} in their DrQA system that used TF-IDF for retrieval and LSTM for the reader which extracts the answer. BERT-based models are the current state-of-the-art \cite{devlin2018} for implementing a reader. A similar approach was used in the physics QA system described in the previous section \cite{zylich2020}. In Generative QA, generative models such as T5 \cite{roberts2020} are used to generate an answer given a question as input. Various contexts can be retrieved and then passed together with the question as an input for the model to use to generate an answer. This technique is called retrieval augmented generation (RAG) \cite{lewis2020}.

In our QA system, we only use the retrieval component to obtain passages that are then displayed to the user as answers. We do not use a reader for our context of answering content-based science questions which could consist of definitions and explanations. This is because short answers such as phrases are generally not adequate but rather, whole contexts that could span multiple continuous sentences. Hence, extracting an answer span is not necessary and could even be inadequate and more error-prone. This point is also made by Zylich et al. \cite{zylich2020} after assessing the misclassifications in their evaluation using machine reading to answer physics course questions. We as a result focus on providing whole passages (a combination of 3 sentences) as answers. Generative QA was not used given that these models were not good at generating answers at the time of this work. It is important to note that this work was done and deployed before the release of ChatGPT and GPT 3.5 API \cite{GPT}, a much better generative model, hence, no comparison can be made to it (see future work section about plans to integrate it). 

\section{Overview of Kwame for Science}
\label{sec:overview}
\begin{figure}
\includegraphics[width=\linewidth]{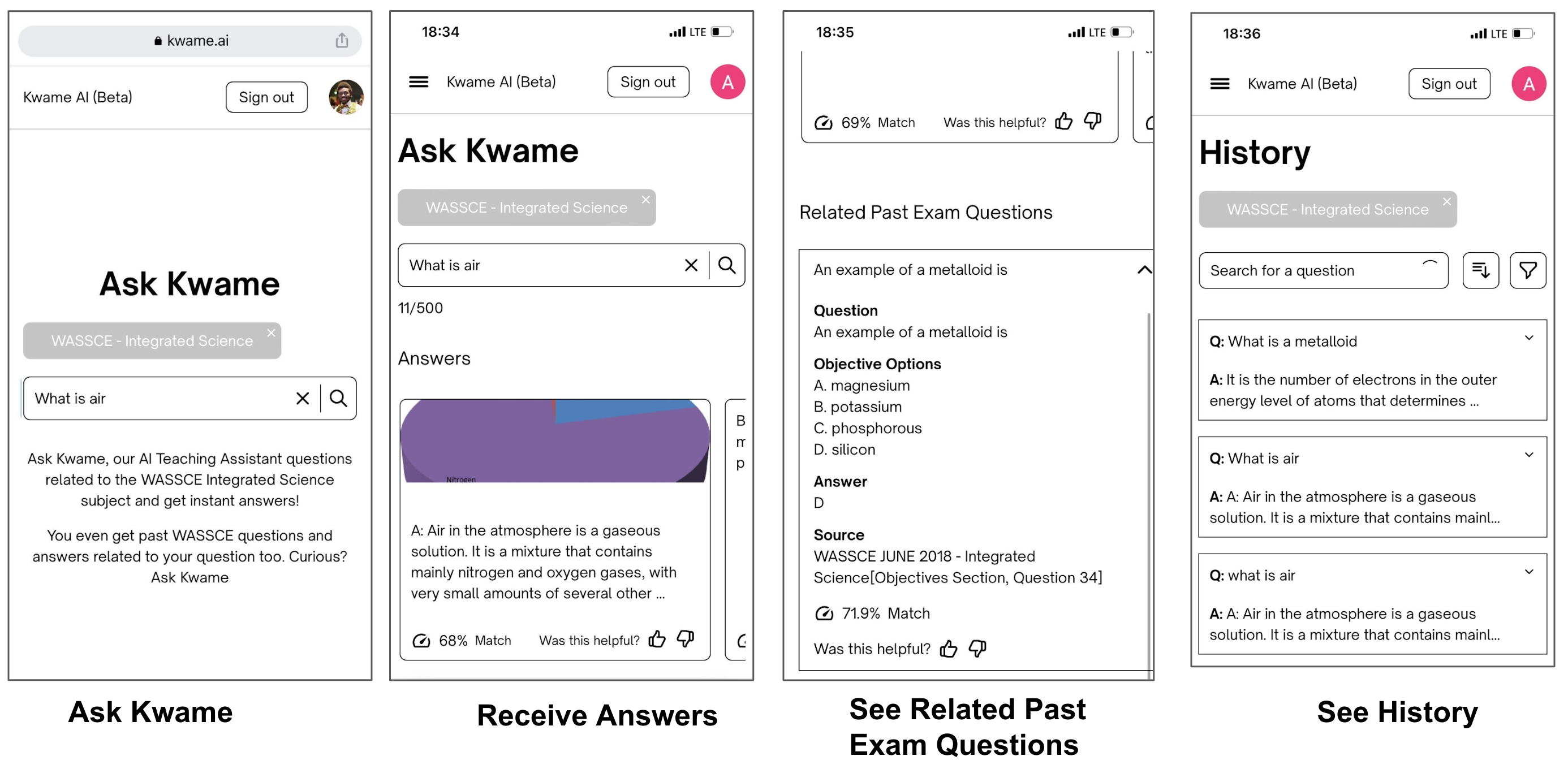}
\caption{Screenshots of QA feature of Kwame for Science} 
\label{fig:kwame4science}
\end{figure}

Kwame for Science is a web app that consists of 2 key features: question answering and viewing past national exam questions. The question-answering (QA) feature enables students to ask science questions and receive 3 passages as answers along with a confidence score per answer which represents the similarity score  (Figure \ref{fig:kwame4science}). In addition to the 3 passages, it displays the top 5 related past national exam questions of the Integrated Science subject and their expert answers. We picked these numbers to ensure we provided several answers (more than 1) without being overwhelming. Also, students can view the history of questions they have asked. The View Past Questions feature enables students to view past national exam questions and answers of the Integrated Science subject and filter the questions based on criteria such as the year of the examination, the specific exam, the type of questions, and topics that were automatically categorized by a topic detection model which we developed. We describe data collection and preprocessing, the QA and view past questions features and the topic detection system.

\subsection{Data Collection and Preprocessing}
Given that our goal was for Kwame to provide answers based on the Integrated Science subject of the WASSCE exam, our training data and knowledge source had to cover the topics in the WASSCE Integrated Science curriculum. We sought to use one of the approved textbooks in Ghana. Unfortunately, their copyrights did not permit such use and the publishers were unwilling to partner with us. Consequently, we searched for free and open-source books and datasets that fulfilled our needs. We came across a U.S. middle school science dataset — Textbook Questions Answering (TQA) \cite{kembhavi2017} which was curated from the free and open-source textbook, CK-12. Our exploration of the dataset revealed that though it covered several of the WASSCE Integrated Science topics, it lacked others, particularly those topics that were contextual such as agricultural science. Consequently, we additionally used a dataset based on Simple Wikipedia to cover those gaps. We used Simple Wikipedia since its explanations use basic words and shorter sentences, and hence, better suited for middle school and high school students compared to regular Wikipedia \cite{SimpleWikipedia}. We parsed the JSON files of the dataset into paragraphs as originally in the dataset. We also extracted figures that were referenced in the paragraphs so they could be returned to students along with the answers. We then split the paragraphs into groups of 3 sentences — passages — using the library sentence-splitter   \footnote{\url{https://pypi.org/project/sentence-splitter/}} resulting in 527K passages. We computed embeddings of the passages using SBERT and indexed them in ElasticSearch because it is open source and scalable to enable fast retrieval and runtime. These constituted the answers returned in response to questions. 

Furthermore, we augmented our question-answering system by adding content specific to the curriculum. In particular, we created question-answer pairs using WASSCE questions of the Integrated Science subject that covered exams from 1993 to 2021 (except 2010 when the exam did not happen). We obtained physical copies of the exams as PDFs, scanned them, and developed software to automatically parse and extract the content using object character recognition (OCR) technology. Given the imperfection of current OCR technology, and in particular, the limitation of extracting mathematical and chemical symbols and equations present in scientific text, we hired and paid individuals to manually inspect and correct all 28 years of exam questions.  We then hired and paid 4 subject-matter experts to provide answers to the exam questions because they did not originally have answers. The experts were Ghanaians and included past contestants of Ghana’s National Science and Math Quiz competition and Senior High School teachers. The experts were given an annotation guidance document on how best to provide their answers, add diagrams, draw tables, and input special characters, among others to ensure that everything, particularly scientific and mathematical symbols, will be rendered properly on the web application. Furthermore, we had initial training rounds where they provided answers for some sample questions and received feedback to ensure consistency. The experts did all the answering online using Google sheets. Diagrams and images which were included in the answers were drawn by experts using third-party tools or they were obtained as open-source images found online. This process results in 3.5K question-answer pairs across 28 years of exams. After obtaining all the answers from the experts, we computed embeddings of the questions similar to the passages of the knowledge base, using SBERT, and indexed them in ElasticSearch. These constituted the related past questions (with answers) that were returned when a question was asked.

\subsection{Question Answering Feature}
\begin{figure}
\includegraphics[width=0.4\linewidth, center]{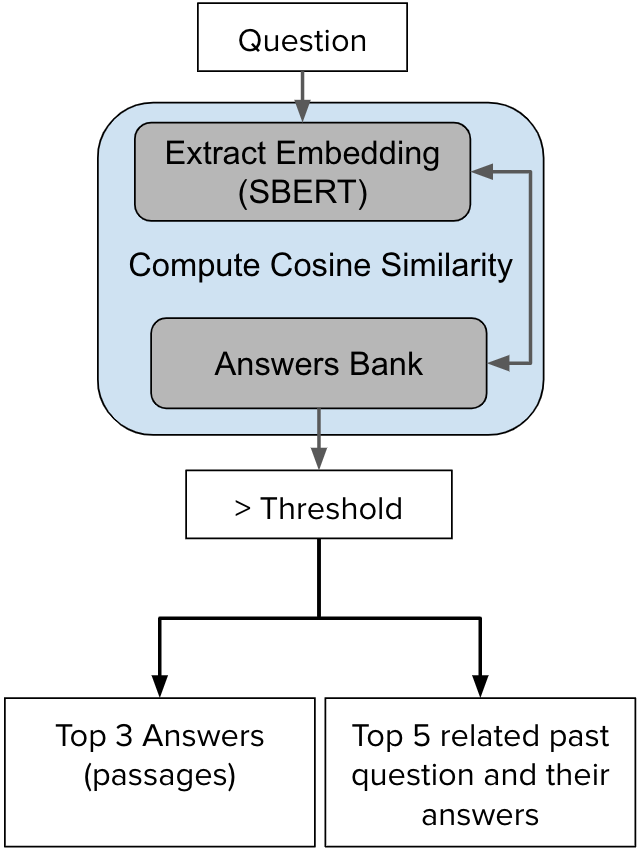}
\caption{Architecture of QA feature of Kwame for Science} 
\label{fig:qa_architecture}
\end{figure}

The question-answering (QA) feature enables students to ask science questions and receive 3 passages as answers along with a confidence score per answer which represents the similarity score  (Figure \ref{fig:qa_architecture}). In addition to the 3 passages, it displays the top 5 related past national exam questions of the Integrated Science subject and their expert answers. When a user types a question in the web app, our system computes an embedding of the question using an SBERT model that was pretrained on 215 M question-answer pairs from diverse sources. We used the SBERT model without fine-tuning it since exploratory evaluation for our science use case which entailed outputting answers for a handful of questions showed it provided adequate answers to the questions. Next, it computes cosine similarity scores with a bank of passages (which are the precomputed embeddings of passages from our knowledge source), retrieves, and returns the top 3 passages along with confidence scores to the web app and any figures or images referenced in that passage which was stored as part of passage’s metadata. Additionally, it computes cosine similarity scores of the question embedding with the precomputed embeddings of the bank of past exam questions, retrieves, and returns the top 5 related questions and their expert answers, along with confidence scores. The web app then displays the answers and the related past exam questions that are above a preset similarity score threshold. If no answer is above the threshold, a message is shown saying the question could not be answered using the knowledge source of that subject. We used precomputed embeddings for fast real-time retrieval which were saved as indices in an ElasticSearch instance we hosted on the Google Cloud Platform. Users could click the answer card to see the passage highlighted in a more detailed answer which is the paragraph. Similarly, for related past question cards, they could click to open them to see the expert answer and other relevant information such as the year of the exam and the question type.

\subsection{View Past Questions}
The View Past Questions feature provides direct access to our data bank of past national exam questions and answers of the Integrated Science subject. Furthermore, students can filter the questions based on criteria such as the year of the examination, the specific exam, the type of questions, and by topic. All these criteria could be inferred easily from the metadata of the original exam files except the topic for which we developed a model to automatically categorize each question according to one of the syllabus topics (see Section \ref{sec:topic_detection}). Utilizing a combination of markdown and LaTeX, we were able to properly render both questions and answers even for niche subjects that make use of equations and symbols not common to regular alphanumeric input, i.e., subjects such as Mathematics or Chemistry. Figures linked to questions or answers are also displayed alongside the question or answer.  Similar to the QA feature, users could click the question cards to open them to see the expert answer and other relevant information such as the year of the exam and the question type.

\subsection{Topic Detection Development}
\label{sec:topic_detection}
We trained a machine-learning topic detection model which we used to classify each of the past exam questions into one of the 48 topics in the Integrated Science subject syllabus. We parsed the official syllabus document which contained a mapping of topics to learning objectives and their corresponding content heading, content details to be covered, example teaching activities, and descriptions of evaluations. The format for these was phrases, sentences, and paragraphs which we call passages. We created a topic-passage pair for each of these mappings resulting in 928 topic-passage samples which we used as the dataset for training and testing. For each passage, we extracted TF-IDF vectors using unigrams and bigrams (baseline approach) and SBERT embeddings (main approach) as features. We then passed the features to a linear support vector machine (SVM) model. We split the data into an 80:20 train-test stratified split. We trained the model on the training set using 5-fold cross-validation and performed hyperparameter turning to pick the best parameters. We then evaluated the trained model on the test set using unweighted average recall (UAR) given that there was a class imbalance. Our baseline approach had a UAR of  83.1\% and our main approach had a UAR of 91\%. The performance is good considering that it is a multi-class classification task with 48 classes showing that our topic detection system is adequate for use. We then trained the SVM model on the whole dataset. Next, we computed embeddings for all past exam questions using SBERT and then performed classification using the trained SVM model to get the topics. These were then imported into the web app as part of the view past questions feature to enable learners to see exam questions for each selected topic.


\section{Real-World Deployment and Evaluation of Kwame for Science}
\label{sec:evaluation}
We launched the web app in beta on 10th June 2022 which was advertised on social media. The statistics for the deployment between 10th June 2022 and 19th February 2023 (8 months) are 750 users across 32 countries (15 in Africa) and 1.5K questions asked.

\subsection	{Question Answering Accuracy}
Users could provide feedback by upvoting or downvoting answers in response to the question “Was this helpful?”. To evaluate Kwame for Science, we used the metrics top 1 and top 3 accuracies. Top 1 accuracy quantifies performance assuming only one answer was returned and voted on. Top 3 accuracy refers to the performance where for each question that received a vote, at least one answer was rated as helpful out of the 3 answers that were returned. Our evaluation results showed 72.6\%  top 1 accuracy (n=197 answers), and 87.2\%  top 3 accuracy (n=109 questions). The top 3 accuracy result is good, showing that Kwame for Science has a high chance of giving at least one useful answer among the 3. We also performed qualitative analysis categorizing the reasons the model incorrectly answered questions to provide insights for future improvements (Table \ref{tab:incorrect_reasons}). Some challenging cases occurred when there were typos in the spelling of scientific words, the questions were incomprehensible and the questions were outside the scope of the knowledge source. Also, some unhelpful answers were cases where the returned paragraph was incomplete due to issues with the dataset, and the answers were related to the question but did not directly provide an answer. These issues are similar to those encountered by Raamadhurai et al. in their real-world deployment of Curio SmartChat \cite{raamadhurai2019}.

\begin{table}[ht]
\centering
\caption{Reasons for unhelpful answers with examples}
\label{tab:incorrect_reasons}
\begin{tabular}{ | p{8em} | p{8em} | p{18em} |}
\hline
\textbf{Reasons} & \textbf{Question} & \textbf{Unhelpful Answer} \\ \hline
Typos in the spelling of scientific words & what is mashine learning & Learning is getting new things into your brain. \\ \hline
Outside the scope of the knowledge source & How is rheumatoid arthritis diagnosed & Rheumatoid arthritis is caused by autoimmunity. Autoimmunity is when the body's immune system attacks its own tissues. The most common places for this arthritis is in the wrists and the knuckles. It usually starts between the ages of 20 and 40. Children and elderly people can also get this kind of arthritis. There are medicines to slow down the process of this disease. There are also some medicines to relieve pain. It is important to treat this kind of arthritis early. \\ \hline
Truncated paragraph & List three excretory organs and their respective products & Examples of exocrine organs and glands: \\ \hline
Related to the question but does not directly provide an answer & Why is the sky blue & Blue is the color of the Earth's sky and sea. Earth looks blue when seen from outer space by astronauts. \\ \hline
Incomplete / incomprehensible queries & water and glass & Light bends when it passes from air to water. The bending of the light causes the pencil to appear broken. Why do waves bend as they enter a new medium \\ \hline
\end{tabular}%
\end{table}

\subsection{Usage Analytics of QA and View Past Questions}
We evaluated the usage of the QA features — clicking to see detailed answers and expanding related questions — and view past questions features — filtering questions by various options and clicking to show an answer  (Table \ref{tab:usage_analytics}). In total, the detailed view of answers was opened 2173 times (1.4 times per question asked). In contrast, the related past questions were expanded 1219 times (0.8 times per question asked). For the viewing past exam questions features, users most frequently used the filtering by year feature (237 times).  This was followed by filtering by question type (174 times) and filtering by topic (104 times). Users could also access detailed answers to past questions by clicking to show the answer and this feature was used 931 times.

\begin{table}[ht]
\centering
\caption{Count for Usage of QA and View Past Questions features}
\label{tab:usage_analytics}
\begin{tabular}{|lc|}
\hline
\multicolumn{1}{|l|}{\textbf{Events}} & \textbf{Count} \\ \hline
\multicolumn{2}{|l|}{\textbf{QA Feature}} \\ \hline
\multicolumn{1}{|l|}{Answer Detail Opened} & 2173 \\ \hline
\multicolumn{1}{|l|}{Related Question Expanded} & 1219 \\ \hline
\multicolumn{2}{|l|}{\textbf{}} \\ \hline
\multicolumn{2}{|l|}{\textbf{View Past Questions Feature}} \\ \hline
\multicolumn{1}{|l|}{Show Answer} & 931 \\ \hline
\multicolumn{1}{|l|}{Select Year} & 237 \\ \hline
\multicolumn{1}{|l|}{Select Question Type} & 174 \\ \hline
\multicolumn{1}{|l|}{Select Topic} & 104 \\ \hline
\end{tabular}
\end{table}

\section{Challenges and Lessons}
\label{sec:challenges}
One of our biggest challenges was getting access to local textbooks to use as a knowledge source for Kwame for Science because of copyright concerns. Despite our relentless effort to get buy-in from local publishers, their disinterest in partnering with us because of trust issues made it impossible for us to power Kwame for Science with content from local textbooks. The trust issue relates to their fear of the softcopy version of their books being redistributed publicly because of the prevalence of such behavior in Ghana. We however addressed this issue by hiring experts to provide answers to past national exam questions to augment our foreign knowledge base with contextual knowledge in the form of expert answers from local experts. This lesson about copyright issues is important for others who might be interested in creating localized educational AI products which will require huge amounts of local data which comes at a huge time and financial cost.

Another issue we had to deal with was the formatting of scientific and mathematical symbols and equations. Unfortunately, the current open-source OCR technologies we tried did not work well for extracting these symbols. Hence, manual effort was needed as an extra quality control step. This issue is important to note especially for developing countries like Ghana and other African countries where important educational documents are mainly accessible in hardcopy format. More work is needed to develop systems that can easily convert scanned scientific documents into outputs that have the correct representation that will render properly in some standard format like markdown. Until then, significant time and monetary effort would be required to generate high-quality data for use.

Another challenge was the difficulty in getting upvotes or downvotes on responses to questions. Even though we had 1.5K questions, we received reactions to only 7.2\% of those. We did not want to force users to provide a rating each time they viewed an answer as that could be annoying. Nonetheless, after the learnings from this deployment and evaluation, we decided to implement the forced rating feature to enable us to obtain a lot of data to evaluate our models in real-time and finetune them. We plan to explore approaches to nudge users rather than force them to provide ratings.

\section{Limitations and Future Work}
\label{sec:limitations}
Our work did not evaluate the effect of Kwame for Science on learning outcomes. In the future, we will deploy Kwame for Science in a controlled setting (specific sets of schools) and run a randomized controlled trial to assess the impact of Kwame for Science on learning outcomes evaluated via end-of-term examinations or national exams.

We did not perform real-time classification of the topic when a question was asked. This information could be useful for generating real-time analytics of the distribution of the topics of the asked questions. Our future work will integrate the topic detection model for real-time classification. An important addition for this to work well is a pre-classification model that classifies the question as a science or non-science topic given users could ask non-scientific questions.

Furthermore, our future work will classify questions into different levels of Bloom's Taxonomy. This information will be useful to understand the types of questions students ask, e.g., definition-type questions that are in lower levels of Bloom’s Taxonomy vs application-type questions in a higher level of Bloom's Taxonomy which is the ultimate goal when facilitating students’ learning.

We will improve the performance of our QA system. We will use the data we collected from our real-world deployment and past exam question-answer pairs to finetune the SBERT model to improve the model’s accuracy. Also, we will include a re-ranking step where we will use a BERT model and cross-attention between pairs of questions and top answers to re-rank the relevant passages returned as answers to improve the accuracy of the entire QA system. Also, we will add generative models like GPT-4 \cite{GPT} to build a retrieval augmented generation (RAG) system \cite{lewis2020} to ensure that our response directly answers users’ questions. 

In the future, we would explore making Kwame for Science available to African students via more accessible channels such as WhatsApp, USSD, toll-free calling and also providing answers in local languages.

\section{Conclusion}
\label{sec:conclusion}
In this work, we developed, deployed, and evaluated Kwame for Science which provides passages from well-curated knowledge sources and related past national exam questions as answers to questions from students. It also enables them to view past national exam questions along with their answers and filter by year, question type, and topics that were automatically categorized by a topic detection model which we developed (91\% UAR). Our deployment of Kwame for Science in the real world over 8 months had 750 users across 32 countries (15 in Africa) and 1.5K questions asked. Our evaluation showed an 87.2\% top 3 accuracy (n=109 questions) implying that Kwame for Science has a high chance of giving at least one useful answer among the 3  displayed. Our usage analytics showed that the detailed view of answer cards was opened 1.4 times per question asked. Also, our categorization of the reasons for incorrectly answered questions included typos in scientific words, incomprehensible questions, incomplete passages, and limited scope of our knowledge source. Some challenges included copyright issues that affected access to local textbooks, inadequate OCR technology for parsing scanned documents, and limited user ratings. Nonetheless, with a first-of-its-kind tool within the African context, Kwame for Science has the potential to enable the delivery of scalable, cost-effective, and quality remote education to millions of people across Africa.

\bibliographystyle{splncs04}
\bibliography{refs}

\begin{thebibliography}{10}
\providecommand{\url}[1]{\texttt{#1}}
\providecommand{\urlprefix}{URL }
\providecommand{\doi}[1]{https://doi.org/#1}

\bibitem{benedetto2019}
Benedetto, L., Cremonesi, P.: Rexy, a configurable application for building virtual teaching assistants. In: IFIP Conference on Human-Computer Interaction. pp. 233--241. Springer (2019)

\bibitem{boateng2021b}
Boateng, G.: Kwame: a bilingual ai teaching assistant for online suacode courses. In: International Conference on Artificial Intelligence in Education. pp. 93--97. Springer (2021)

\bibitem{boateng2021}
Boateng, G., Annor, P.S., Kumbol, V.W.A.: Suacode africa: Teaching coding online to africans using smartphones. In: Proceedings of the 10th Computer Science Education Research Conference. pp. 14--20 (2021)

\bibitem{boateng2022}
Boateng, G., John, S., Glago, A., Boateng, S., Kumbol, V.: Kwame for science: An ai teaching assistant based on sentence-bert for science education in west africa. In: Proceedings of the Fourth International Workshop on Intelligent Textbooks (iTextbooks) 2022 co-located with 23d International Conference on Artificial Intelligence in Education (AIED 2022). pp. 26--30 (2022)

\bibitem{boateng2019}
Boateng, G., Kumbol, V.W.A., Annor, P.S.: Keep calm and code on your phone: A pilot of suacode, an online smartphone-based coding course. In: Proceedings of the 8th Computer Science Education Research Conference. pp. 9--14 (2019)

\bibitem{chen2017}
Chen, D., Fisch, A., Weston, J., Bordes, A.: Reading wikipedia to answer open-domain questions. arXiv preprint arXiv:1704.00051  (2017)

\bibitem{devlin2018}
Devlin, J., Chang, M.W., Lee, K., Toutanova, K.: Bert: Pre-training of deep bidirectional transformers for language understanding. arXiv preprint arXiv:1810.04805  (2018)

\bibitem{goel2020}
Goel, A.: Ai-powered learning: Making education accessible, affordable, and achievable. arXiv preprint arXiv:2006.01908  (2020)

\bibitem{goel2016}
Goel, A.K., Polepeddi, L.: Jill watson: A virtual teaching assistant for online education. Tech. rep., Georgia Institute of Technology (2016)

\bibitem{GPT}
Chat completions api. \href{https://platform.openai.com/docs/guides/text-generation/chat-completions-api}{https://platform.openai.com/docs/guides/text-generation/chat-completions-api}

\bibitem{kembhavi2017}
Kembhavi, A., Seo, M., Schwenk, D., Choi, J., Farhadi, A., Hajishirzi, H.: Are you smarter than a sixth grader? textbook question answering for multimodal machine comprehension. In: Proceedings of the IEEE Conference on Computer Vision and Pattern recognition. pp. 4999--5007 (2017)

\bibitem{lewis2020}
Lewis, P., Perez, E., Piktus, A., Petroni, F., Karpukhin, V., Goyal, N., K{\"u}ttler, H., Lewis, M., Yih, W.t., Rockt{\"a}schel, T., et~al.: Retrieval-augmented generation for knowledge-intensive nlp tasks. Advances in Neural Information Processing Systems  \textbf{33},  9459--9474 (2020)

\bibitem{raamadhurai2019}
Raamadhurai, S., Baker, R., Poduval, V.: Curio smartchat: a system for natural language question answering for self-paced k-12 learning. In: Proceedings of the Fourteenth Workshop on Innovative Use of NLP for Building Educational Applications. pp. 336--342 (2019)

\bibitem{rajpurkar2016}
Rajpurkar, P., Zhang, J., Lopyrev, K., Liang, P.: Squad: 100,000+ questions for machine comprehension of text. arXiv preprint arXiv:1606.05250  (2016)

\bibitem{reimers2020}
Reimers, N., Gurevych, I.: Making monolingual sentence embeddings multilingual using knowledge distillation. In: Proceedings of the 2020 Conference on Empirical Methods in Natural Language Processing. Association for Computational Linguistics (11 2020), \url{https://arxiv.org/abs/2004.09813}

\bibitem{roberts2020}
Roberts, A., Raffel, C., Shazeer, N.: How much knowledge can you pack into the parameters of a language model? In: Proceedings of the 2020 Conference on Empirical Methods in Natural Language Processing (EMNLP). pp. 5418--5426 (2020)

\bibitem{SimpleWikipedia}
Simple english wikipedia. \href{https://simple.wikipedia.org/wiki/Main\_Page}{https://simple.wikipedia.org/wiki/Main\_Page}

\bibitem{UNESCO2020}
Unesco. pupil-teacher ratio sub-saharan africa. \href{https://data.worldbank.org/indicator/SE.PRM.ENRL.TC.ZS?locations=ZG}{https://data.worldbank.org/indicator/SE.PRM.ENRL.TC.ZS?locations=ZG} (Feb 2020)

\bibitem{zylich2020}
Zylich, B., Viola, A., Toggerson, B., Al-Hariri, L., Lan, A.: Exploring automated question answering methods for teaching assistance. In: International Conference on Artificial Intelligence in Education. pp. 610--622. Springer (2020)

\end{thebibliography}


\end{document}